\pdfoutput=1

\documentclass[3p,times,procedia]{elsarticle}

\usepackage{ecrc}
\usepackage[bookmarks=false]{hyperref}
    \hypersetup{colorlinks,
      linkcolor=blue,
      citecolor=blue,
      urlcolor=blue}

\usepackage[utf8]{inputenc}
\usepackage[T1]{fontenc}

\volume{00}

\firstpage{1}

\journalname{Procedia Computer Science}

\runauth{Lujun Li et al.}

\jid{procs}

\usepackage{amssymb}
\usepackage{adjustbox}
\usepackage{booktabs}
\usepackage{caption}
\usepackage{multirow}
\usepackage{svg}

\usepackage[figuresright]{rotating}

\begin{document}
\begin{frontmatter}



\dochead{International Neural Network Society Workshop on Deep
Learning Innovations and Applications}

\title{Exploring the Impact of Temperature on Large Language Models: Hot or Cold?}


\author[a]{Lujun Li\corref{cor1}} 
\author[a]{Lama Sleem}
\author[b]{Niccolo' Gentile}
\author[b]{Geoffrey Nichil}
\author[a]{Radu State}

\address[a]{University of Luxembourg}
\address[b]{Foyer S.A.}

\begin{abstract}
The sampling temperature, a critical hyperparameter in large language models (LLMs), modifies the logits before the softmax layer, thereby reshaping the distribution of output tokens. Recent studies have challenged the "Stochastic Parrots" analogy by demonstrating that LLMs are capable of understanding semantics rather than merely memorizing data and that randomness, modulated by sampling temperature, plays a crucial role in model inference. In this study, we systematically evaluated the impact of temperature in the range of 0 to 2 on data sets designed to assess six different capabilities, conducting statistical analyses on open source models of three different sizes. Small (1B-4B), medium (6B-13B), and large (40B-80B). Our findings reveal distinct skill-specific effects of temperature on model performance, highlighting the complexity of optimal temperature selection in practical applications. To address this challenge, we propose a BERT-based temperature selector that takes advantage of these observed effects to identify the optimal temperature for a given prompt. We demonstrate that this approach can significantly improve the performance of small and medium models in the SuperGLUE datasets. Furthermore, our study extends to FP16 precision inference, revealing that temperature effects are consistent with those observed in 4-bit quantized models. By evaluating temperature effects up to 4.0 in three quantized models, we find that the ``Mutation Temperature''—the point at which significant performance changes occur - increases with model size\footnote{\url{https://github.com/DobricLilujun/temperature_eval}}. 
\end{abstract}

\cortext[cor1]{Corresponding author. Tel.: +0033-766636416.}
\end{frontmatter}

\email{lilujun588588@gmail.com}

\vspace*{-6pt}

\section{Introduction}
\label{introduction}

Since the release of ChatGPT, LLMs have significantly impacted both academia and industry, revolutionizing the development of artificial intelligence. Open source models of different sizes have facilitated advances in various domains, \cite{zhao2023survey} including question answering and summarization. A key factor influencing the performance of LLMs is hyperparameter adjustment. For example, Top-K sampling selects the next token from the $K$ most probable candidates, while Top-P sampling samples from the smallest set of tokens whose cumulative probability exceeds $P$ \cite{fan-etal-2018-hierarchical}. Additionally, the repetition penalty reduces the probability of tokens that have already appeared, helping to avoid repetition. In this paper, we focus on temperature, which is one of the most frequently used hyperparameters. During inference in LLMs, this parameter is used to scale the logits of the output layer, effectively controlling the randomness of model predictions. The concept of temperature, denoted as $T$ \cite{ACKLEY1985147}, was first introduced by Ackley, who emphasized its crucial role in shaping the Boltzmann distribution. Formally, the probability $P_i$ of the $i$-th token is given by $P_i = \frac{e^{y_i / T}}{\sum_{j=1}^{V} e^{y_j / T}}$, where $y_i$ denotes the pre-softmax activation of the $i$-th token (commonly called the logit), $T$ represents the temperature, and $V$ is the total number of tokens in the vocabulary.
The value $P_i$ determines the probability that the $i$-th token will be generated, after which the model output is produced by a sampling algorithm. As $T$ increases, the probability mass function (PMF)~\cite{chang2023kldivergence} becomes more uniform; conversely, as $T$ approaches zero, the distribution collapses to a delta function, causing the algorithm to behave greedily by always selecting the most likely token. At each generation step, a new token is selected by randomly sampling the updated probability distribution \cite{DBLP:conf/iclr/HoltzmanBDFC20}.



In this study, we focus on three key research questions (RQs): (RQ1) \textbf{To what extent does temperature impact the performance of LLMs across different abilities?} (RQ2) \textbf{Does temperature have uniform effects across different abilities and models, and what are the main differences observed?} (RQ3) \textbf{Is there an optimal temperature for each capability, and can the best temperature be determined for a specific prompt?} The remainder of this paper is organized as follows: Section~2 reviews related work. Section~3 describes the experimental methods. Section~4 details the experimental settings. Section~5 presents and analyzes the results. Finally, Section~6 concludes the paper.

\section{Related Work}
\label{sec:Performance analysis}

Investigations of the effects of temperature remain limited in the recent literature. Most studies report results using only one temperature value, without systematically exploring a wider range, except for the series of Llama models, which tested two settings for code generation\cite{touvron2023llama}. Furthermore, general data sets designed to evaluate multiple model capabilities simultaneously, such as those for Artificial General Intelligence (AGI; \cite{zhong-etal-2024-agieval}), tend to lack a specific focus on a single foundational ability. For example, higher temperatures increase creativity~\cite{10.1609/aaai.v38i1.27798}, while lower temperatures improve logical reasoning. For tasks like complex math problems that need both logic and creativity, these effects may cancel each other out. This makes it difficult to see the real impact of temperature, a phenomenon we call the ``Temperature Paradox''.

Recent studies have examined how temperature settings affect different tasks and dynamic configurations. \cite{Renze_2024} explored temperature in multitask scenarios using prompt engineering and a range from 0 to 1, finding no significant effect on LLM performance. \cite{DBLP:conf/icccrea/PeeperkornK0J24} investigated temperature in creative writing, measuring perplexity and cosine similarity, and found only a weak effect on creativity. \cite{10.1609/aaai.v38i1.27798} proposed an adaptive temperature strategy for code generation, assigning higher temperatures to harder tokens (such as the start of a Python function) and lower temperatures to tokens with greater model confidence, showing that higher temperatures can help with complex tasks. However, there is still no clear guideline for choosing temperature for different LLMs, tasks, or prompts, although temperature adjustment is important for LLM users, RAG systems, and agentic AI systems.

\section{Approaches}

To address the ``Temperature Paradox'' and more accurately measure the effect of temperature on each ability with minimal bias, we adopt datasets with clear capability preferences and employ a single-prompt format, querying the model only once to avoid multi-prompt assistance. This approach enables a more precise and unbiased assessment of the intrinsic abilities of LLMs. We hypothesize that temperature influences different model abilities in distinct ways. Therefore, our study focuses on six core intrinsic abilities that not only represent the primary competencies of LLMs but are also central to computational linguistics research.

\subsection{Evaluating Intrinsic Abilities}
\label{Intrinsic ability}

\textit{Causal Reasoning (CR)}: A cognitive faculty historically ascribed solely to humans that consists in deriving conclusions from given premises by adhering to strict logical principles\cite{kiciman2024causal}. In this paper, we use CRASS \cite{frohberg-binder-2022-crass}, a publicly available counterfactual reasoning data set that simplifies the evaluation process by requiring the model to select the correct answer rather than generate it. \textbf{Top-1 Accuracy (T1)} is used to quantify the frequency with which the model correctly predicts the true class by selecting the class with the highest confidence after multiple repetitions.
    
\textit{Creativity (CT)}: Creativity involves the generation of novel and valuable ideas, concepts, or products that require both originality and effectiveness \cite{article123456}. For CT, we adopt a framework that assesses stories in four dimensions: fluency, flexibility, originality, and elaboration, using customized questions based on the Torrance Test of Creative Writing (TTCW) procedure \cite{10.1145/3613904.3642731}. In this framework, each category includes multiple standard evaluation questions, with true or false determined by expert judgment. Twelve publicly available New Yorker stories are used as plots. The \textbf{TTCW Accuracy} is then calculated by counting the number of positive evaluations among all Q\&A pairs \cite{GUZIK2023100065}.

    
\textit{In-Context Learning (ICL)}: ICL reflects an LLM's ability to understand text and perform tasks using contextual information and a few examples \cite{Radford2018ImprovingLU}. In this study, we focus on the LongBench-TREC long-context task \cite{bai-etal-2024-longbench}, where the model learns from a sequence of questions and answers and must classify a final question based on this context. \textbf{Classification Score (CLS)}, measured by accuracy, evaluates the model's ability to recognize patterns and make correct predictions compared to the ground truth.

\textit{Instruction Following (IF)}: IF measures the model's ability to follow instructions provided in the prompts, which is essential for effective LLM applications. For this study, we used InfoBench \cite{qin-etal-2024-infobench}, which introduces the Decomposed Requirements Following Ratio (\textbf{DRFR}) as a metric to assess the performance of the follow-up of instruction. DRFR decomposes complex instructions for more granular evaluation and has demonstrated greater reliability and effectiveness.


\textit{Machine Translation (MT)}: MT evaluates an LLM's ability to translate text between languages, a key area where LLMs have shown strong performance. We use the FLORES-101 benchmark \cite{goyal-etal-2022-flores} for multilingual evaluation, adopting \textbf{spBLEU} as the metric. BLEU scores measure the similarity between model outputs and reference translations. To ensure comparability, we normalize the spBLEU scores by dividing by 100, so that all results fall within the range [0, 1]. Given the prevalence of English in the LLM training data, we focus on English-to-other-language translation, selecting diverse pairs (e.g., English-to-Maltesian, Indonesian, Latvian, Icelandic and Khmer) to cover varying levels of translation difficulty.

\textit{Summarization (SUMM)}: Summarization aims to condense long texts into concise summaries while preserving key information and main ideas. One of the main challenges in this task is the reliable evaluation. To address this, we use the ``benchmark\_llm\_summarization'' dataset \cite{zhang-etal-2024-benchmarking}, which provides expert-written reference summaries. For evaluation, we adopt the reference-based metric \textbf{Rouge-L F1}, which measures the overlap of the longest common subsequence between generated and reference summaries, balancing precision and recall, and has been shown to correlate well with human judgments \cite{lin-och-2004-automatic}.

\subsection{LLM-as-a-Judge Evaluation}

\begin{figure*}[htbp]
    \centering
    \begin{minipage}[!htbp]{0.53\textwidth}
        \captionsetup{justification=raggedright,singlelinecheck=false}
        \captionof{table}{\centering{Selected datasets and evaluation metrics}}
        \label{tab:table dataset selection}
        \centering
        \begin{adjustbox}{max width=1.0\textwidth}
            \begin{tabular}{l l l l l l}
                \toprule
                Ability & Dataset & Samples & Metrics & Source & Evaluations\\
                \midrule
                \textbf{CR} & CRASS & 3500 & Top-1 accuracy & \cite{frohberg-binder-2022-crass}  & GPT3.5\\
                \textbf{CT} & Creativity\_eval & 84 & TTCW Accuracy &  \cite{10.1145/3613904.3642731} & GPT3.5 \\
                \textbf{ICL} & LongBench-TREC & 1015 &  CLS score & \cite{bai-etal-2024-longbench}  & Exact Matching\\
                \textbf{IF} & InfoBench & 3500 & DRFR &  \cite{qin-etal-2024-infobench} & GPT3.5 \\
                \textbf{MT} & Flore101 & 2100 & Normalized spBLEU & \cite{goyal-etal-2022-flores} & SPM tokenizers \\
                \textbf{SUMM} &  benchmark\_llm\_summarization & 2114 & Rouge-L F1 Score  &\cite{zhang-etal-2024-benchmarking} &  Exact Matching \\
                \bottomrule
            \end{tabular}
        \end{adjustbox}
    \end{minipage}%
    \hfill
    \begin{minipage}[!htbp]{0.43\textwidth}
        \centering
        \includegraphics[width=0.95\textwidth,height=0.65\textheight,keepaspectratio]{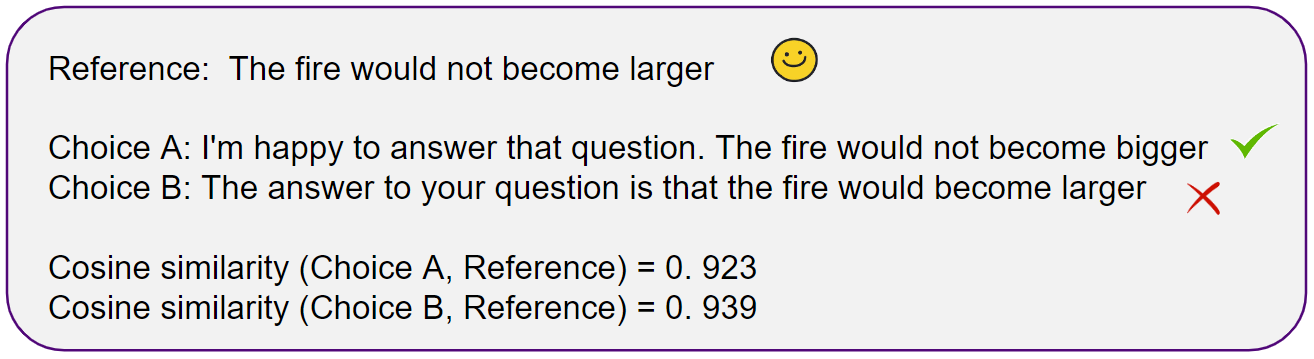}
        \captionof{figure}{Cosine Similarity Using BERT Embedding Model}
        \label{fig:img-example-cosine-simularity}
    \end{minipage}
\end{figure*}

LLM-as-a-Judge has been widely used and has been proven to be highly aligned with human judgment \cite{liu-etal-2023-g}. Due to the inherent stochasticity of LLMs and the flexibility in textual expressions that convey identical meanings, particularly in small and medium models, unintended results often appear preceding or following the target response. This makes reference-based evaluations, such as exact matching or similarity metrics, particularly challenging. For example, similarity metrics have two main issues: (1) It is hard to set a clear cutoff for correct answers; (2) Embedding-based methods often miss key words like ``not,'' as shown in Figure~\ref{fig:img-example-cosine-simularity}~\cite{anschutz-etal-2023-correct}.

Advanced models such as GPT-4o and DeepSeek~\cite{10.5555/3495724.3495883,bi2024deepseek} have achieved strong results on many benchmarks. For tasks with complex answers or challenging evaluation, we use LLM-as-a-Judge. For CR, CT, and IF, we use ChatGPT with carefully designed prompts instead of exact matching or human annotation. These judgments are used to calculate the metrics described in Section~\ref{Intrinsic ability}. For the other three abilities with simple reference answers, we use standard evaluation methods, as shown in Table~\ref{tab:table dataset selection}.




\section{Experiments}

\subsection{General Experiment Settings}

In this study, experiments begin with the selection of diverse benchmark datasets designed to challenge state-of-the-art (SOTA) models, as shown in Table~\ref{tab:models}. Evaluation is carried out primarily in a question-answer format or through matching functions, with specific metrics applied to each dataset. All models are quantized to 4 bits using the AWQ method \cite{10.1145/3714983.3714987}, and vLLM \cite{kwon2023efficient} is used as the default inference acceleration framework. Each question is tested three times across 12 models. 

\begin{table}[ht]
\centering
\begin{adjustbox}{max width=0.90\textwidth}
\begin{tabular}{p{4cm}c@{\hskip 1em}c|p{4cm}c@{\hskip 1em}c|p{4cm}c@{\hskip 1em}c}
\toprule
\multicolumn{3}{c|}{\textbf{Small Size Models (1B - 4B)}} & \multicolumn{3}{c|}{\textbf{Medium Size Models (6B - 13B)}} & \multicolumn{3}{c}{\textbf{Large Size Models (40B - 80B)}} \\
\addlinespace

\textbf{Model} & \textbf{Size} & \textbf{Date} & \textbf{Model} & \textbf{Size} & \textbf{Date} & \textbf{Model} & \textbf{Size} & \textbf{Date} \\
\midrule
Llama-3.2-1B-Instruct    & 1.2B & Sep 2024 & Llama-2-7b-chat-hf       & 6.7B  & Jul 2023 & Llama-2-70b-chat-hf        & 69.0B & Jul 2023 \\
Llama-3.2-3B-Instruct    & 3.2B & Sep 2024 & Llama-2-13b-chat-hf      & 13.0B & Jul 2023 & Meta-Llama-3-70B-Instruct  & 70.6B & Apr 2024 \\
Phi-3.5-mini-instruct    & 3.8B & Jun 2024 & Mistral-7B-Instruct-v0.2 & 7.2B  & Mar 2024 & Mixtral-8x7B-Instruct-v0.1 & 46.7B & Dec 2023 \\
Qwen2.5-1.5B-Instruct    & 1.5B & Sep 2025 & Meta-Llama-3-8B-Instruct & 8.0B  & Apr 2024 & --                         & --    & --       \\
Qwen2.5-3B-Instruct      & 3.1B & Sep 2025 & --                       & --    & --       & --                         & --    & --       \\
\bottomrule
\end{tabular}
\end{adjustbox}
\caption{Investigated Small, Medium, and Large models with their respective sizes and release dates.}
\label{tab:models}
\end{table}

The temperature settings range from 0.1 to 1.9 in increments of 0.3, resulting in seven distinct configurations for each model. Temperatures above 2.0 are excluded, as previous research has shown that higher values tend to produce non-informative and excessively incoherent text \cite{DBLP:journals/corr/abs-1811-02549}. Each model is evaluated using only one question per test. We consistently used gpt-3.5-turbo-0125 as the evaluation model with a temperature setting of 0.01, and selected open source models as listed in Table~\ref{tab:models}. Nucleus sampling was adopted, as it yields perplexity values closest to human text \cite{DBLP:conf/iclr/HoltzmanBDFC20}, with the following parameters: max\_length = 4096, Top\_P = 0.9, repetition penalty (RP) = 1.0, and max\_new\_tokens = 1024.


\subsection{Supplementary  Experiment}

\subsubsection{Best Temperature Selection On SuperGLUE}
\label{sec:superglue}

SuperGLUE is a benchmark consisting of eight tasks for evaluating word-sense disambiguation, natural language inference, coreference resolution, and question answering \cite{NEURIPS2019_4496bf24}. The main results from the previous section can be used to identify the optimal temperature for a given prompt and model, provided that the primary ability required for the prompt is known. To this end, a classification model based on a fine-tuned BERT framework, denoted as ``BERT-based Selector", is proposed, which is trained on the experimental prompts tailored for the main experiments and will finally be used in classifying every input prompt. Another option is to obtain the required ability for the prompt based on the well-designed prompt that will be asked to GPT models, denoted as ``GPT-based Selector".

The basic idea of this selector is the following: Let $F$ be a selection model, either BERT or GPT, acting as an ability predictor based on a given prompt $x_p$. The model output $F(x_p) $ represents the predicted ability most closely associated with the given prompt $x_p$. For example, given a training prompt such as $x_p$ = \textit{``Translate `J'aime le chat' from French to English''}, the model $F(x_p)$ would be expected to predict ``MT'' (Machine Translation) as the most required ability. As such, the optimal temperature parameter \( T^* = \arg\max_{T} \mathcal{D}(T, F(x_p), M) \) is selected to maximize the estimated performance of the model \(M\) on the task \(F(x_p)\) under different temperature settings, where \(\mathcal{D}()\) represents the performance distribution of a given model \(M\) over temperatures obtained from previous main experimental results.

To test this framework, we evaluated three models of different sizes—Llama-3.2-1B-Instruct (Small), Llama-3-8B-Instruct (Medium), and Mixtral-8x7B-Instruct-v0.1 (Large)~\cite{NEURIPS2019_4496bf24}—on the SuperGLUE benchmark. Each question in the benchmark was asked three times, and each instance was generated by incrementing the random seed, initially set to 42, by 1 for each successive iteration.

\subsubsection{Experiments On Extended Settings}

We also investigated the temperature range $[0, 4]$ while maintaining 4-bit quantization. Although we observed performance degradation and inconsistent generations at temperatures above 2, we did not find evidence for a specific ``Mutation Temperature'' in large models. To further explore the effect of inference precision, we repeated the main experiments on the same three models using FP16 precision, focusing on the temperature range from 0 to 2, to determine whether temperature effects differ when inference precision is altered. Additionally, since Top-$K$ and Top-$P$ sampling influence the output distribution at the candidate selection level, while RP operates at the logits level, we further evaluated various settings for Top-$K$ (2, 5, 10), Top-$P$ (0.8, 0.9, 1.0), and RP (0.0, 1.0, 2.0). These experiments were conducted on all three models mentioned above to systematically assess the impact of these parameters on performance.

\section{Results and Analysis}
\subsection{Findings from Statistical Analysis}

Table~\ref{table:Statistical_Metrics} provides a summary of the temperature-performance correlations in six abilities, based on results from three categories of models. ``P. Coef.'' and ``S. Coef.'' correspond to the Pearson and Spearman correlation coefficients, respectively. ``Range (\%)Max'' may be interpreted as the relative performance variation across temperatures, while ``Range Max (\%)'' refers to the maximum relative ranges within the size category. ``CV'' (Coefficient of variation) represents the ratio of average performance to standard deviation, and ``CV Max'' indicates the highest CVs observed within the size category. The average accuracy and standard deviation for the temperatures and models within each group are also reported. 

\begin{table*}[!htbp]
\centering
\caption{Comparison of temperature-performance correlations for six abilities across three model categories.}
\label{table:Statistical_Metrics}
\begin{adjustbox}{max width=0.8\textwidth}
\begin{tabular}{rcccccccccccccccc}
\toprule
\multirow{2}{*}{\textbf{Ability}} & 
\multirow{2}{*}{\textbf{P. Coef.}} & 
\multirow{2}{*}{\textbf{P. p-value}} & 
\multirow{2}{*}{\textbf{S. Coef.}} & 
\multirow{2}{*}{\textbf{S. p-value}}   & \multicolumn{3}{c}{\textbf{Range Max (\%)}}        & \multicolumn{3}{c}{\textbf{CV Max}}              & \multicolumn{3}{c}{\textbf{Average Accuracy}}  & \multicolumn{3}{c}{\textbf{Standard Deviation}} \\

\cmidrule(lr){6-8}
\cmidrule(lr){9-11}
\cmidrule(lr){12-14}
\cmidrule(lr){15-17}

\multicolumn{1}{l}{} & \multicolumn{2}{l}{} & \multicolumn{2}{l}{} & Small               & Medium               & Large              & Small              & Medium              & Large              & Small             & Medium             & Large              & Small              & Medium              & Large             \\

\midrule
\textbf{CR}          & -0.07       & 0.00        & -0.07    & 0.00           & 146.02          & 49.37           & 19.41          & 58.79          & 14.88          & 6.43           & 0.41          & 0.52          & \textbf{0.82}  & 0.05           & 0.03           & \textbf{0.02} \\
\textbf{CT}          & \textbf{-0.14}  & 0.00    & -0.10   & 0.00            & 186.81          & 154.55          & \textbf{82.02} & \textbf{82.64} & \textbf{72.90} & \textbf{28.07} & 0.36          & \textbf{0.45} & \textbf{0.47}  & \textbf{0.27}  & 0.12           & \textbf{0.08} \\
\textbf{ICL}         & -0.10     &      0.00    & -0.09      & 0.00         & 122.04          & 55.52           & 20.19          & 48.83          & 21.66          & 7.20           & 0.38          & 0.26          & \textbf{0.49}  & 0.06           & 0.04           & \textbf{0.01} \\

\textbf{IF}          & \textbf{-0.40}   &      0.00   & \textbf{-0.37}    &      0.00  & 154.65          & 116.63          & 22.03          & 72.22          & 47.64          & 8.04           & 0.49          & 0.68          & \textbf{0.73}  & 0.26           & 0.08           & \textbf{0.02} \\

\textbf{MT}          & \textbf{-0.216}   &      0.00   & -0.40      &      0.00         & \textbf{192.32} & \textbf{162.59} & 76.86          & \textbf{91.09} & \textbf{72.14} & \textbf{27.35} & \textbf{4.72} & 5.95          & \textbf{11.55} & 3.19           & \textbf{2.54}  & \textbf{1.96} \\
\textbf{SUMM}        & \textbf{-0.51}     &      0.00  & -0.45        &      0.01       & 154.29          & 89.20           & 4.35           & 72.89          & 32.70          & 1.57           & 0.16          & 0.21          & \textbf{0.23}  & 0.09           & 0.02           & \textbf{0.00}  \\

\bottomrule
\end{tabular}
\end{adjustbox}
\end{table*}

In this table, it can be observed that the performance of IF, MT, and SUMM exhibits relatively strong correlations with temperature, as indicated by both correlation coefficients. The statistical significance of these correlations is further supported by p-values of zero. Furthermore, both ``Range Max'' and ``CV Max'' decrease as the size of the model increases, which statistically suggests that larger models are more robust to temperature-induced variations. The average accuracy metric further demonstrates that larger models achieve higher statistical performance across all six abilities. In particular, performance differences among models of different sizes are relatively small for CT, IF, and SUMM, but much more pronounced for CR, ICL, and MT. These findings provide practical guidance for selecting the model size according to specific functional requirements.

\subsection{Temperature Effects on Different Abilities}



\begin{figure}[!h]
\centering
    \includegraphics[width=0.8\textwidth]{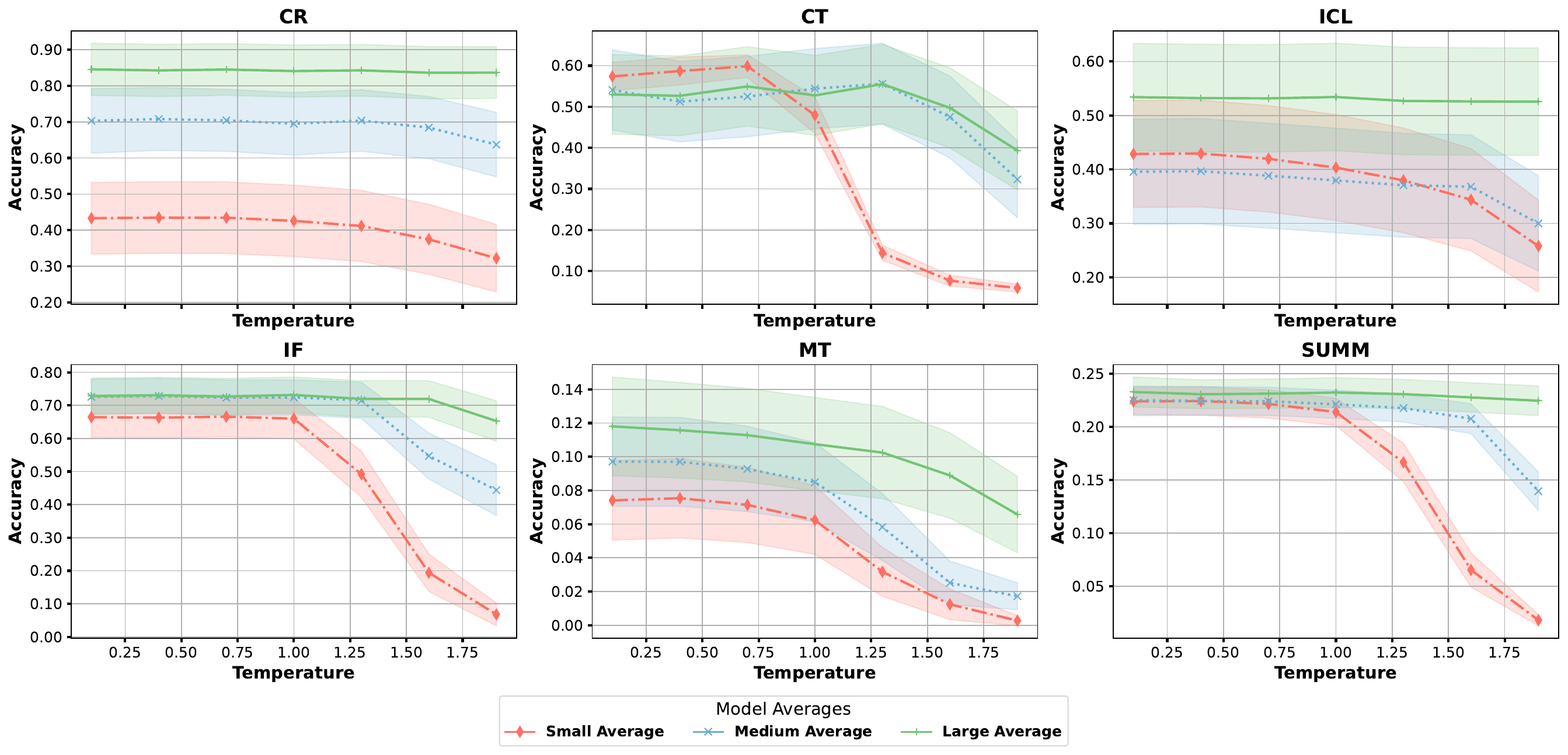}
    \caption{Average performance trends for different model sizes, with shaded bands indicating variability.}
    \label{fig:Experiment_Results_category_size}
\end{figure}

Figure~\ref{fig:Experiment_Results_category_size} illustrates the impact of temperature on models of varying sizes across a range of evaluated abilities. Lines show the mean performance for each model size, while shaded regions correspond to $\pm 0.2$ standard deviations. For the sake of consistency, all evaluation metrics enumerated in Table~\ref{tab:table dataset selection}---including spBLEU for machine translation---are uniformly referred to as ``accuracy'' and have been normalized to the interval $[0, 1]$.

\paragraph{Causal Reasoning (CR)}

CR questions are counterintuitive and require logical reasoning, each with three options. Medium and large models exceed the 33.3\% random baseline, while small models perform only slightly above this chance level—by approximately 7\%—across most temperature settings, indicating limited reasoning ability. The large and medium models show slight improvement at a temperature of 1.3, suggesting that higher temperatures may help to address complex problems. The optimal temperature for CR is not always zero and an increase in temperature does not necessarily reduce performance. In contrast, small models do not demonstrate substantial causal reasoning ability within the scope of this study.

\paragraph{Creativity (CT)}

An optimal temperature of 1.3 is recommended for medium and large models to maximize creativity. Small models show a marked decline in creativity at $T=1.0$, while medium and large models are only affected at $T=1.7$. Generally, temperature first increases and then decreases creativity. Small models are more creative at lower temperatures, but large models are more robust and generate more diverse outputs, as indicated by their wider shaded regions. In general, temperature strongly influences creativity, with moderate values being the most beneficial.

\paragraph{In-Context Learning (ICL)}

Large models achieve the best average performance, while the difference between medium and small models is minimal. This indicates that ICL, as an emerging property of LLMs, requires a sufficiently large model size, highlighting the significance of scaling laws. Medium models show less performance decline than small models. At a temperature of 1.7, small models degrade faster than medium models, despite outperforming them at lower temperatures. Large models maintain stable performance across temperatures from 0 to 2, with no abrupt performance drop observed. Increasing temperature generally reduces performance, although large models may experience slight improvements at higher temperatures.

\paragraph{Instruction Following (IF)}

The behavior of IF is particularly noteworthy. As the temperature increases from 0 to 1, IF performance remains largely unchanged. However, when the temperature exceeds 1, different models experience relatively pronounced negative effects, and the larger the model size, the later these negative effects emerge. Performance changes with temperature are abrupt: small models exhibit a mutation between 1.0 and 1.3, medium models between 1.3 and 1.6, and large models demonstrate a moderate mutation temperature from 1.6 to 1.9. Therefore, for users of LLMs who require strict adherence to instructions, it is advisable to set the temperature below 1.

\paragraph{Machine Translation (MT)}

Slightly increasing the temperature within the low range marginally improves translation performance for small and medium models only. The rise in temperature has the most detrimental effect on MT, as indicated by the highest range of performance and CV in Table~\ref{table:Statistical_Metrics}. This trend can be attributed to the inherently deterministic nature of translation, and all models exhibit comparable declines in performance. The optimal temperature is close to zero ($0 + \epsilon$), and language understanding performance depends primarily on the breadth of the training data and the model's parameter size.

\paragraph{Summarization (SUMM)}

Temperature effect curves are initially stable but drop sharply at higher temperatures, especially for small models. Statistical analysis shows a strong negative correlation between performance and temperature. SUMM tasks follow a similar trend to IF tasks, but the mutation temperature for medium models is higher (about $1.7$), and large models show no clear mutation temperature.


\subsection{Supplementary Experiment}

\subsubsection{Best Temperature Selection on SuperGLUE}

We conducted experiments on three models, as shown in Table~\ref{tab:superglue}. The table presents the SuperGLUE validation accuracy under different temperature settings: $ACC_D$ denotes the accuracy with the \textbf{D}efault temperature of 1.0, while $ACC_B$ and $ACC_C$ represent the precision achieved by dynamically selecting the optimal temperature using our fine-tuned \textbf{B}ERT model and \textbf{C}hatGPT-based prompting, respectively. This comparison clearly demonstrates the performance difference from optimal temperature selection. 

\begin{table}[!htbp]
\centering
\caption{SuperGLUE validation accuracy under default and dynamically selected temperature settings.}
\label{tab:superglue}
\begin{adjustbox}{max width=0.40\textwidth}
\begin{tabular}{cccccc}
\toprule
\textbf{Model}                                       & \textbf{Type} & \textbf{COPA} & \textbf{WIC} & \textbf{WSC} & \textbf{Average} \\
\midrule
\multirow{3}{*}{\textbf{Llama-3.2-1B-Instruct}}      & $ACC_D$        & 0.510         & 0.196        & 0.346        & 0.252            \\
 & $ACC_B$ & 0.600 & 0.500 & 0.356 & \textbf{0.494} \\
 &  $ACC_C$ & 0.600 & 0.477 & 0.365 & 0.477 \\ 
 \midrule
\multirow{3}{*}{\textbf{Meta-Llama-3-8B-Instruct}}   & $ACC_D$        & 0.860         & 0.547        & 0.673        & 0.600            \\
 & $ACC_B$ & 0.900 & 0.556 & 0.673 & \textbf{0.612} \\
 & $ACC_C$ & 0.900 & 0.549 & 0.664 & 0.605 \\
 \midrule
\multirow{3}{*}{\textbf{Mixtral-8x7B-Instruct-v0.1}} & $ACC_D$        & 0.800         & 0.608        & 0.298        & 0.593            \\
 & $ACC_B$ & 0.800 & 0.608 & 0.298 & 0.593 \\
 & $ACC_C$ & 0.800 & 0.608 & 0.298 & 0.593 \\
 \bottomrule
\end{tabular}
\end{adjustbox}
\end{table}

Adjusting the temperature can greatly improve the performance of WIC (one of the tasks in SuperGLUE) for Llama-3.2-1B and Meta-Llama-3-8B-Instruct. This shows that the optimal temperature selector provides stable performance, avoiding potential performance drops that can occur when using a fixed temperature. When working with small models, this is indeed one of the necessary parameters to consider, especially in resource-constrained scenarios. Considering that SuperGLUE primarily evaluates a range of different capabilities, our optimal temperature selector still demonstrates consistent improvements. It is important to mention that this selector does not inherently boost performance—Supervised Fine-Tuning (SFT) remains the primary method—but it ensures that the model achieves the best possible performance by avoiding suboptimal settings. For large models, we did not observe significant performance differences, indicating that optimizing the temperature setting is generally less critical for larger models. However, as suggested in previous findings, when large models are used to solve complex reasoning tasks, a higher temperature can sometimes lead to performance gains. Therefore, adjusting the temperature may still be necessary in such scenarios.



\subsubsection{Results on Extended Settings}

\begin{figure}[!htbp]
\centering
  \includegraphics[width=0.8\columnwidth]{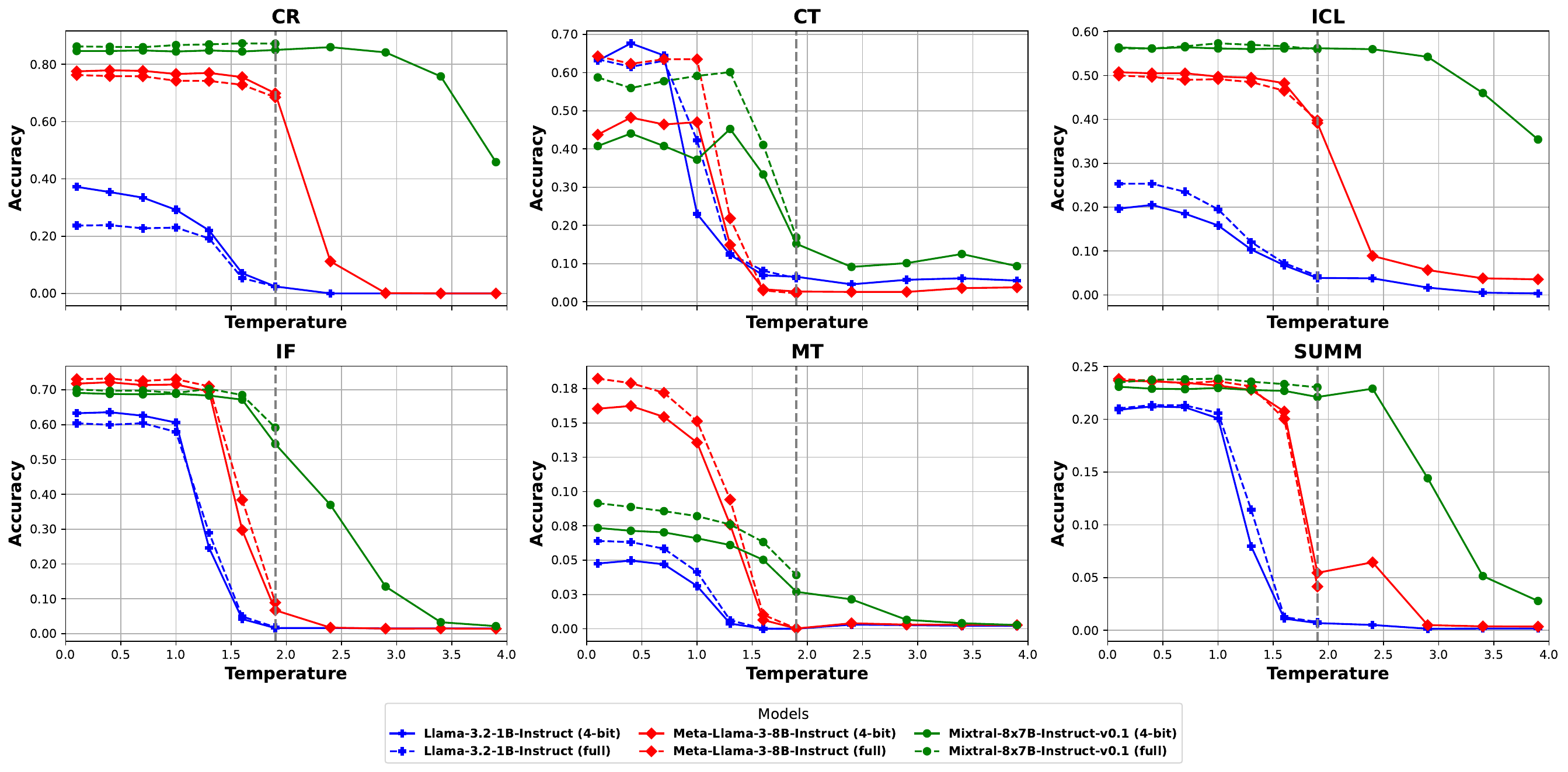}
  \caption{Performance with extended temperature to 4.0}
  \label{fig:Performance_with_extended_temperature_settings_and_inference_precision}
\end{figure}

\paragraph{Extension to 4.0} 
Fig.~\ref{fig:Performance_with_extended_temperature_settings_and_inference_precision} presents the performance curves of models with 4-bit precision across six capabilities as temperature varies; ``Full'' refers to FP16 precision inference. Extending the temperature range helps identify both the ``mutation temperature,'' where performance drops sharply, and the potential upper limit for temperature settings. Large models are more robust to increasing temperatures, while small models experience significant performance loss at lower temperatures, which aligns with our expectations. In particular, each model has its own mutation temperature, with larger models generally exhibiting a higher threshold.


\paragraph{FP16 Precision} 

Fig.~\ref{fig:Performance_with_extended_temperature_settings_and_inference_precision} also shows that the optimal temperature of the models does not exhibit significant changes across the entire temperature range. Overall, the results indicate that there is no substantial difference in optimal temperature between the two levels of precision, although there is a 10\%--20\% difference in performance. Since our main experiments were conducted under 4-bit quantization, these findings also validate the effectiveness of our results and demonstrate their scalability with respect to inference precision.

\paragraph{Other Parameters}

\begin{table}[!htbp]
\centering
\resizebox{0.98\textwidth}{!}{%
\begin{tabular}{lcccccccccccccccccc}
\toprule
\multirow{2}{*}{\textbf{Experiment}} & \multicolumn{3}{c}{\textbf{CR}} & \multicolumn{3}{c}{\textbf{CT}} & \multicolumn{3}{c}{\textbf{ICL}} & \multicolumn{3}{c}{\textbf{IF}} & \multicolumn{3}{c}{\textbf{MT}} & \multicolumn{3}{c}{\textbf{SUMM}} \\
\cmidrule(lr){2-4} \cmidrule(lr){5-7} \cmidrule(lr){8-10} \cmidrule(lr){11-13} \cmidrule(lr){14-16} \cmidrule(lr){17-19}

& $p_1$ & $p_2$ & $p_3$ & $p_1$ & $p_2$ & $p_3$ & $p_1$ & $p_2$ & $p_3$ & $p_1$ & $p_2$ & $p_3$ & $p_1$ & $p_2$ & $p_3$ & $p_1$ & $p_2$ & $p_3$ \\
\midrule
\textbf{(Top-P, RP) = (0.9, 1.0) ; Top-K $\in$ \{2, 5, 10\}} & 1.00 & 0.99 & 1.00 & \textbf{0.27} & \textbf{0.46} & \textbf{0.57} & 1.00 & 0.99 & 0.99 & 0.90 & 0.86 & 0.99 & 0.98 & 0.98 & 0.99 & 0.86 & 0.83 & 0.99 \\
\textbf{(Top-K, RP) = (5, 1.0)   ; Top-P = [0.8, 0.9, 1.0]} & 1.00 & 0.99 & 1.00 & \textbf{0.53} & \textbf{-0.29} & \textbf{-0.33} & 0.98 & 0.96 & 0.98 & 0.98 & 0.91 & 0.91 & 0.98 & 0.99 & 0.99 & 0.97 & 0.97 & 0.99 \\
\textbf{(Top-P, Top-K) = (0.9, 5)  ; RP = [0.0,1.0,2.0]}     & 1.00 & 1.00 & 1.00 & \textbf{0.18} & \textbf{-0.08} & \textbf{0.47} & 1.00 & 1.00 & 1.00 & 0.86 & \textbf{0.58} & 0.90 & 0.99 & 0.99 & 1.00 & 0.98 & 0.99 & 0.99 \\
\bottomrule
\end{tabular}
}
\caption{Pairwise Pearson correlation coefficients for performance curves with varying parameter values.}
\label{tab:topk_topp}
\end{table}

We analyze the effect of varying a single parameter by computing the Pearson correlation coefficients between the resulting performance curves, while keeping all other parameters fixed. Specifically, for each parameter, we select three distinct values and calculate the pairwise Pearson correlation coefficients among their corresponding temperature-performance curves. The values $p_1$, $p_2$, and $p_3$ denote the correlations for the first vs. second, first vs. third, and second vs. third parameter settings, respectively. These results are summarized in Table~\ref{tab:topk_topp}. It can be clearly observed that, except for CT, the correlation of temperature effects between different parameter settings is very high, indicating minimal impact. In contrast, RP has a significant effect on CT and IF. For CT, the impact of Top-$K$ is smaller than that of RP, while Top-$P$ exerts the greatest influence, as indicated by the relatively low or even negative correlation coefficients. Therefore, when the model is required to perform creative tasks, it is essential to set Top-$P$ appropriately, and RP should also be configured with care. However, these three parameters exert minimal influence on the temperature performance curve and warrant particular consideration only for CT and IF tasks.

\section{Conclusion}

In this study, we extend previous work on temperature effects in language models by examining six capabilities over a wider temperature range. We also improve the GPT evaluation methods and the testing protocol, focusing on both statistical and empirical analysis of temperature effects. We also compare the performance of the model under FP16 and the quantization of 4-bits, finding minimal differences in the temperature effect. By extending the temperature range to 4.0, we observe that large models still have a mutation temperature, but it is higher than 2.0. Additionally, we introduce BERT-based and GPT-based temperature selectors, demonstrating their effectiveness on the SuperGLUE dataset, especially for small models.

To answer RQ1, \textbf{``To what extent does temperature affect the performance of LLMs across multiple abilities?"}, we find that temperature has a modest effect on In-Context Learning and Causal Reasoning, but can significantly impact Machine Translation (up to 192.32\%), Creativity (up to 186.81\%) in small models (see Table~\ref{table:Statistical_Metrics}). Spearman coefficients indicate a generally negative correlation between temperature and performance. 


To answer RQ2, \textbf{``Does temperature exert uniform effects on different abilities across models, and what are the primary differences observed?''}, we find that large models are more resilient to temperature changes. Increasing the temperature slightly improves Causal Reasoning, In-Context Learning, and Instruction Following, followed by a decline in performance. For Summarization and Machine Translation, higher temperatures generally have a negative impact, especially on smaller models. The influence of temperature varies substantially across different abilities and model sizes, making it difficult to generalize a consistent pattern; the optimal temperature also differs significantly between models. Furthermore, high temperatures are not always detrimental to logic-oriented abilities such as Causal Reasoning, Instruction Following, and In-Context Learning.


To answer RQ3, \textbf{``Is there an optimal temperature for each ability, and can the best temperature be found for a specific prompt?''}, we find that no single temperature is optimal for all tasks. Higher temperatures are not always best for creative writing, nor is zero always best for following instruction. However, by identifying the required ability for each prompt using our BERT model and referencing our experimental results, we can select the optimal temperature for each prompt. In three SuperGLUE tasks, this approach improves performance for small and medium-sized models.

This study has some limitations. We examined only the effect of temperature on six text-based ability; additional abilities, such as planning and coding, could also be evaluated in future work. It is difficult to verify the accuracy of the GPT model evaluations without manually checking each case. More testing is needed to determine whether the optimal temperature selector is effective for other models of similar or larger sizes. Furthermore, a mathematical explanation of how temperature affects model performance requires further investigation.

\section*{Acknowledgements}

This research has greatly benefited from the collaboration and support of our industrial partner, Foyer S.A., as well as from academic institutions, particularly the Interdisciplinary Centre for Security, Reliability and Trust (SnT), and numerous individual contributors. We would like to thank the ``AI \& Data Studio'' team for their valuable insights, guidance, and for providing essential computational resources (NVIDIA H100 GPUs), which were crucial for conducting our experiments. The guidance and mentorship provided by faculty and advisors, together with constructive suggestions and technical expertise shared by postdoctoral researchers, have significantly improved the quality and impact of this work. Furthermore, we exclusively used publicly available datasets and models, and we affirm that no AI-generated text was used in the preparation of this manuscript. Throughout the research process, we have adhered to ethical standards and maintained transparency, ensuring that all of our methods and results are clearly communicated.



\appendix

\bibliography{custom.bib}
\bibliographystyle{elsarticle-harv}






\clearpage


\end{document}